\documentclass[10pt,twocolumn,letterpaper]{article}

\usepackage{cvpr}
\usepackage{times}
\usepackage{epsfig}
\usepackage{amsmath}
\usepackage{amssymb}
\usepackage{graphicx} 
\usepackage{algorithm}
\usepackage{algorithmic}
\usepackage{multirow}
\usepackage{booktabs}
\DeclareMathOperator*{\argmin}{argmin}

\usepackage[breaklinks=true,bookmarks=false]{hyperref}

\cvprfinalcopy 

\def\cvprPaperID{****} 

\begin{document}

\title{Furnishing Your Room by What You See: An End-to-End Furniture Set Retrieval Framework with Rich Annotated Benchmark Dataset}

\author{Bingyuan Liu$^1$,~Jiantao Zhang$^1$,~Xiaoting Zhang$^2$,~Wei Zhang$^2$,~Chuanhui Yu$^1$,~and~Yuan~Zhou$^1$\\
        $^1$Kujiale.com, Hangzhou, China $^2$Zhejiang University, Hangzhou, China \\
        $^{1}$\{badi, xiahua, haicheng, zhouyuan\}@qunhemail.com \\ $^{2}$xtzhang@zju.edu.cn, zhangwei1995\_zju@163.com
}

\maketitle

\begin{abstract}
Understanding interior scenes has attracted enormous interest in computer vision community.
However, few works focus on the understanding of furniture within the scenes and a large-scale dataset is also lacked to advance the field.
In this paper, we first fill the gap by presenting DeepFurniture, a richly annotated large indoor scene dataset, including 24k indoor images, 170k furniture instances and 20k unique furniture identities.
On the dataset, we introduce a new benchmark, named furniture set retrieval. Given an indoor photo as input, the task requires to detect all the furniture instances and search a matched set of furniture identities.
To address this challenging task, we propose a feature and context embedding based framework.
It contains $3$ major contributions:
(1) An improved Mask-RCNN model with an additional mask-based classifier is introduced for better utilizing the mask information to relieve the occlusion problems in furniture detection context.
(2) A multi-task style Siamese network is proposed to train the feature embedding model for retrieval, which is composed of a classification subnet supervised by self-clustered pseudo attributes and a verification subnet to estimate whether the input pair is matched.
(3) In order to model the relationship of the furniture entities in an interior design, a context embedding model is employed to re-rank the retrieval results.
Extensive experiments demonstrate the effectiveness of each module and the overall system.
\end{abstract}

\section{Introduction}

Indoor scene understanding has become a popular research topic in recent years due to its significance and challenge in both academic research and industrial applications\cite{xiao2018unified}. 
Most related works focus on the tasks of scene classification\cite{lazebnik2006beyond}, layout estimation\cite{zou2018layoutnet} and scene parsing\cite{zhao2017pspnet}.
However, few works focus on furniture understanding, which is also critical for the ultimate scene understanding and reconstruction.
On the other side, this may be due to the lack of a well annotated large-scale dataset like ImageNet\cite{deng2009imagenet} to motivate the research.
Previous furniture databases are either undersized\cite{badami20173d} or designed for a specific task like categorization or material prediction.
Some scene understanding databases\cite{xiao2018unified}\cite{li2018interiornet} are organized and labeled very well, but they mainly support scene understanding tasks.

\begin{figure}[t]
\centering
\includegraphics[width=1.0\linewidth]{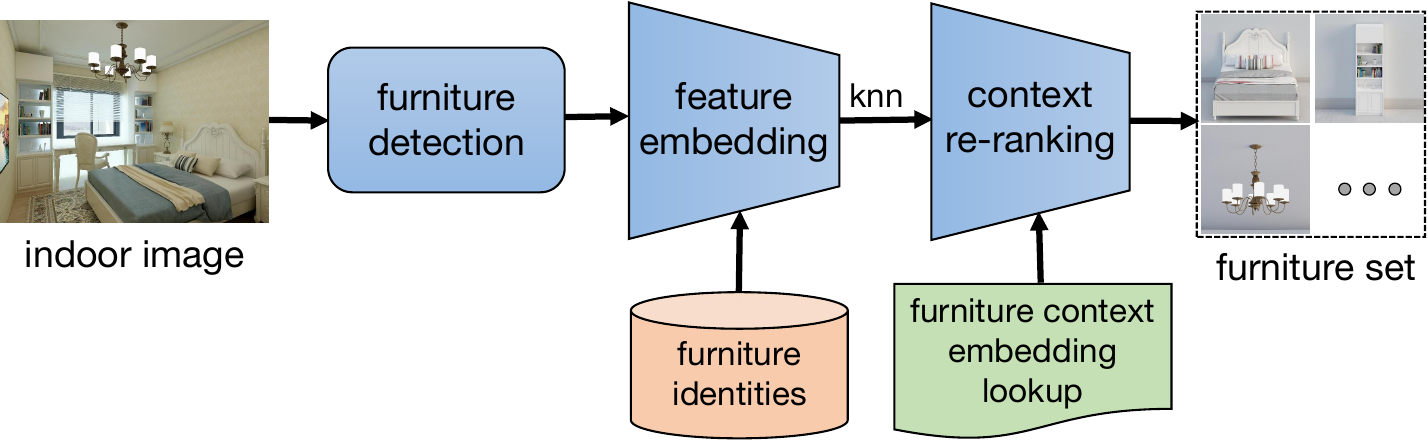}
\caption{The proposed furniture set retrieval framework.}
\label{fig:pipeline}
\end{figure}


\begin{figure*}[htb]
\centering
\includegraphics[width=1.0\linewidth]{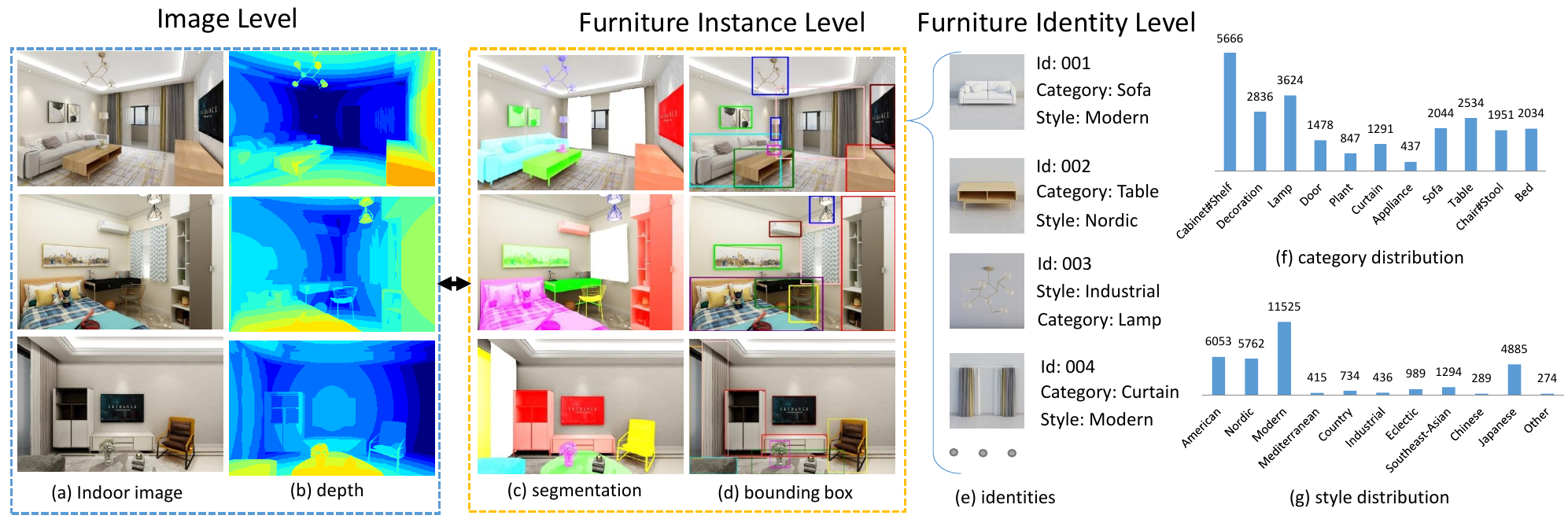}
\caption{The DeepFurniture dataset has hierarchical annotations, \emph{i.e.}, image level, instance level and identity level. (f) and (g) show the category and style distribution of identities in our dataset}
\label{fig:data}
\end{figure*}

For the purpose of furniture understanding, we present a large-scale DeepFurniture dataset.
It includes about 24k indoor images, 170k furniture instances and 20k unique furniture identities.
As shown in Figure \ref{fig:data}, this dataset is richly annotated on three levels, \emph{i.e.}, image level, furniture instance level and furniture identity level.
Thus a full spectrum tasks can be benchmarked on it including categorization, 
detection, segmentation, and retrieval.
Based on its richness and uniqueness, the dataset also introduces a new benchmark, named furniture set retrieval.
Given an indoor photo as input, this task requires to detect all the furniture instances of interest and search a matched set of items from a large-scale furniture identity databse.
The diversity of the dataset makes this task extremely challenging.

To address the furniture set retrieval problem, we propose an end-to-end framework based on feature and context embedding.
As shown in Figure \ref{fig:pipeline}, it consists of three major modules, \emph{i.e.}, furniture detection, instance retrieval based on feature embedding and context re-ranking.
First, an improved Mask-RCNN model is developed to detect the furniture of interest in an image, which better leverages the mask information by adding an classifier in the mask branch to relieve the issues of occlusion in the furniture setting.
Second, we propose a Siamese network for feature embedding by integrating a verification subnet and a classification subnet.
The model is trained in a multi-task fashion, where the verification branch aims to learn an optimized feature for matching input pairs, and the other branch distinguish the input by self-clustered attributes.
Here we utilize the clustering method to obtain a spectrum of attributes as the supervision in classification as it can bring better regularization for feature learning. 
Once the feature embedding model is obtained, furniture retrieval is performed by exhaustive search over feature index of furniture identities.
At last, we train a context embedding model to encode the collaborative relationship of furniture items in an indoor room and use it to re-rank the retrieval results.
Extensive experiments on the DeepFuniture dataset demonstrate the effectiveness of the proposed method compared with baselines.

\section{Related Work}

\textbf{Furniture datasets}. To the best of our knowledge, the computer vision community lacks a large-scale dataset particular built for furniture understanding.
\cite{lim2013parsing} propose a dataset of IKEA 3D models and aligned images, while \cite{badami20173d} introduces a RGBD dataset for furniture model.
However, these two datasets are both undersized.
In \cite{hu2017visual}, a furniture dataset is introduced for the purpose of furniture style analysis.
During recent years, several well organized scene datasets are proposed, such as SUNCG\cite{song2017semantic}, Broaden\cite{bau2017network} and ADE20K\cite{zhou2017scene}.
Different from previous works, our dataset is designed especially for versatile tasks in furniture analysis. 

\textbf{Object detection}. Current object detection methods can be roughly divided into one-stage models and two-stage models.
The former ones like YOLO\cite{redmon2016you}, SSD\cite{liu2016ssd} and RetinaNet\cite{lin2017focal} mainly focus on the efficiency, while the two-stage methods usually achieve better performance, such as Faster R-CNN\cite{fasterrcnn} and Mask R-CNN\cite{he2017mask}.
In this paper, we employ the two-stage framework.
It first proposes candidate object bounding boxes and then performs object classification and location regression.
In Mask R-CNN, a mask branch is introduced into the second stage to employ the segmentation information to co-train the model.
The most related work with our method is Mask Score R-CNN\cite{huang2019mask}, which adds another subnet in mask branch to learn the mask quality.
Some other proposed improvements can also be integrated into the detection framework, such as FPN\cite{fpn}, deformable convolution\cite{zhu2019deformable}, soft-nms\cite{bodla2017soft}, etc.

\textbf{Image retrieval and verification}. Deep learning \cite{krizhevsky2012imagenet}\cite{he2016deep} based methods has dominated the research of image retrieval and person verification, because they greatly improve the feature representation.
It is reported that the classification formulation is also effective for the retrieval task\cite{geng2016deep} \cite{DeepFace2014}, but more improvements are achieved by casting the retrieval task as a deep metric learning problem.
Some training objectives are proposed, such as pairwise verification loss\cite{ahmed2015improved}, contrastive loss\cite{radenovic2018fine} and triplet ranking loss\cite{liu2017end}.
Correspondingly the overall network architecture is a Siamese CNN network with either two or three branches for the pairwise or triplet loss.
In contrast, our model is a Siamese two-branch network with an classification loss and pairwise verification loss respectively.
This architecture is similar to the one used for person re-identification\cite{geng2016deep}.
Different from the situation of person verification, our dataset suffers more serious data imbalance and samples for each identity are rare to some extent.
Thus we incorporate the verification loss and classification loss, which are both effective and complementary.
Another difference is that we employ self-clustered attributes as the supervision in classification loss rather than the identity ID or labeled categories.
This is inspired by the work of deep clustering\cite{caron2018deep} and we demonstrate that only one iteration of clustering is enough to achieve good performance in our task.
Some re-ranking methods are considered to refine the retrieval results like \cite{zhong2017re} \cite{bai2016sparse}, while in our method a context embedding model is employed to collaboratively improve the search results of a furniture set.

\section{DeepFurniture Dataset}
\label{sec3}

To advance the research on furniture understanding, we present a dataset named DeepFurniture.
Figure \ref{fig:data} shows some samples and the annotation hierarchy.
To the best of our knowledge, DeepFurniture is a large-scale furniture database with the richest annotations for a versatile furniture understanding benchmark.

\subsection{Data Collection and labeling} 
All the data samples are contributed by millions of designers and artists on a product-level interior design platform.
On this platform, users can create a computer-aided design (CAD) floor plan, drag and drop 3D models, such as furniture, bricks, wallpapers and so on, to design a room or a house. 
Via the high-quality and high-speed rendering service, they can then generate photo-realistic renderings for visualization.
The platform has already accumulated millions of 3D model and billions of interior designs and images.

To avoid copyright issues and obtain a high quality dataset, we carefully select 24k indoor images from user-generated rendering images.
These images contain more than 170k furniture instances related to 20k furniture identities.
The labels of the dataset are naturally available due to the generation process, but noises exist to a large extent.
Thus, we refer to human labors to check and clean all the annotations like category and style.

The annotations of the dataset are organized into three levels: image, furniture instance and furniture identity, as depicted in Figure \ref{fig:data}.
On image level, each indoor image is attached to one scene category such as living room, dining room, bedroom and study room.
A depth map is also provided along with each image.
On furniture instance level, the bounding boxes and per-pixel masks of the instances in each image are given.
In the furniture identity set, each entity is an actual 3D model and we use one high-quality rendering preview image to represent it in our dataset.
Each identity is referred to as a unique ID and comes with its category label and style tags.
The categories cover $11$ major furniture classes, such as cabinet, table, chair, sofa, etc.
And the style tags are annotated by some professional designers, including $11$ types, such as modern, country, Chinese, industrial and so on.
A brief distribution statistics of the identities is indicated in Figure \ref{fig:data}, and the size of the whole identity set is about 20k.
The number of furniture categories is $11$, and the number of furniture style tags is $11$ as well.
On average, each identity has 6.9 instances in different images with various views and context.

\subsection{Benchmarks}
The rich annotations of DeepFurniture make it a great fit for multiple furniture understanding tasks, \emph{i.e.}, scene categorization, depth estimation, multi-style classification, furniture detection, segmentation and retrieval.
In this paper, we benchmark $3$ major tasks as follows.

\textbf{Furniture detection and segmentation.} Furniture detection task aims to detect furniture instances in a given image by predicting bounding boxes and category labels, while the task of segmentation assigns a category label to each pixel of an instance.
We employ the standard evaluation metrics in COCO\cite{lin2014microsoft}.
For the detection task, we employ the bounding box's average precision $AP_{box}$, $AP_{box}^{IoU=0.50}$ and $AP_{box}^{IoU=0.75}$, where $AP_{box}$ is computed by averaging over IoU thresholds.
Similarly, the metric of segmentation is average precision computed over masks, denoted as $AP_{mask}$, $AP_{mask}^{IoU=0.50}$, and $AP_{mask}^{IoU=0.75}$.

\textbf{Furniture instance retrieval.} Our dataset is quite suitable to the task of furniture instance retrieval.
Given a detected furniture instance in the interior scene as query, the target of this task is to search a matched item in the furniture identity database.
Since we use an image of furniture instance to search the preview images of identity actually, the task can also be considered as a cross-domain retrieval problem.
To benchmark this task we assume ground-truth bounding boxes are provided as we hope to emphasize the retrieval performance.
Top-k retrieval accuracy is employed as the evaluation metric and mean accuracy is averaged over all categories.

\textbf{Furniture set retrieval.} In realistic application, users are willing to obtain a furniture identity set from a photo for their design project.
To formulate the problem, we define a task named furniture set retrieval.
Given an interior image as input, it aims to find out a matched set from the furniture identity database.
To evaluate the accuracy of furniture set retrieval, we utilize a modified retrieval top-k accuracy, where a correct search means the ground truth set is included in the combinations across the top-k results for each furniture item.

\section{Proposed framework} 
\label{sec4}

\subsection{Improved Mask-RCNN for furniture detection}

\begin{figure}[tb]
\centering
\includegraphics[width=1.0\linewidth]{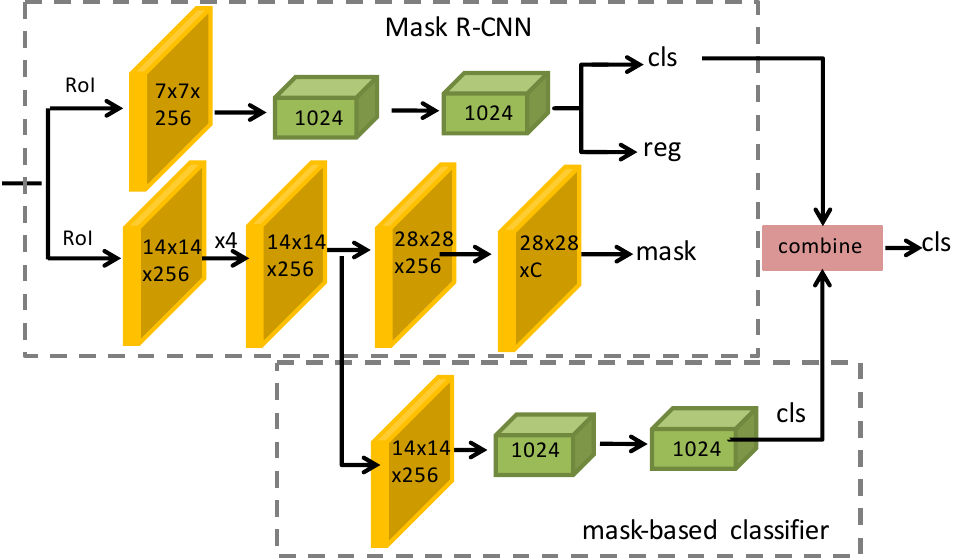}
\caption{Architecture of the improved Mask R-CNN.}
\label{fig:aux_cls}
\end{figure}

The two-stage Mask R-CNN\cite{he2017mask} model has demonstrated that the performance of detection can be enhanced by incorporating a mask branch.
The first stage proposes candidate object bounding boxes regardless of object categories by the RPN network.
Then the second stage performs specific object classification, position regression and mask estimation on the proposals after ROI Align.
Compared to general object detection\cite{lin2014microsoft}, the examples in DeepFurniture suffer more occlusion and confusion issues.
As shown in Figure \ref{fig:data}, the bounding boxes suffer heavy interaction problems and the segmentation may give more guidance to obtain better classification score.

\begin{figure*}[htb]
\centering
\includegraphics[width=1.0\linewidth]{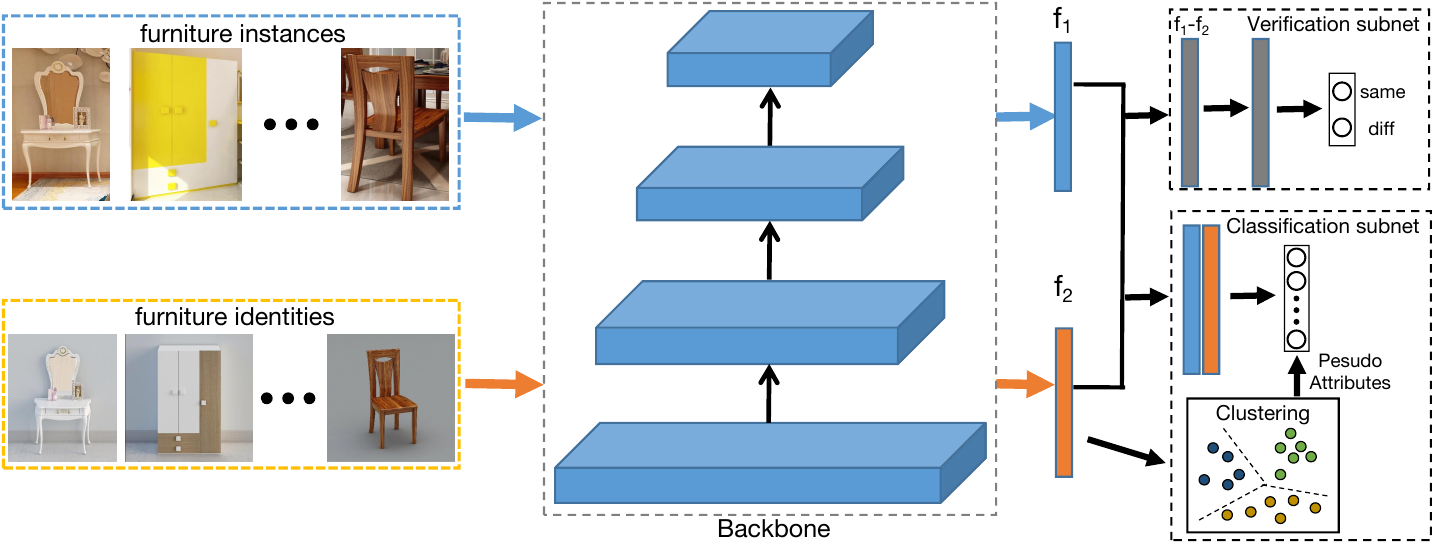}
\caption{The architecture of our Siamese network. A backbone is used to extract feature embedding. Then the verification subnet and the classification subnet are concatenated in a multi-task fashion. }
\label{fig:verify}
\end{figure*}

model by adding an additional mask-based auxiliary classifier.
Figure \ref{fig:aux_cls} presents the main architecture of our model.
It introduces a classifier branch in the mask branch after two convolution layers since it is considered that this layer has encoded enough segmentation information.
The network is also trained in a multi-task method and the loss function is defined as follows:
\begin{equation}
     L = L_{cls}+L_{reg}+L_{mask}+L_{aux\_cls}
\end{equation}
where the first three terms are same as Mask-RCNN and the last one is categorical cross-entropy term introduced by our method. 
Similar to detection branch, a background class is used in the mask-base classifier, and negative samples are also fed into the mask branch. 
During inference, the outputs of the two classifiers are combined as the final detection score.
Two proved effective improvements are also incorporated in our implementation, that is, DCN\cite{zhu2019deformable} and soft-nms\cite{bodla2017soft}.

\subsection{Furniture feature embedding}

Feature embedding aims to find a shared latent space where instances with the same identity close to each other.
As shown in Figure \ref{fig:verify}, we employ a Siamese network by integrating two branches, \emph{i.e.}, pairwise verification subnet and classification subnet.

\subsubsection{Classification subnet}

Some previous works apply item ID or category information as labels to train the feature embedding model\cite{krizhevsky2012imagenet}\cite{geng2016deep}.
However, in our scenario the instances are extremely imbalanced for different entities, while the category information can not give enough visual distinction
Therefore, we refer to a spectrum of attributes generated by clustering.

With the goal of learning the pseudo visual attributes of furniture, a category constrained k-means clustering method is introduced.
First, we extract features of preview images of furniture identities by ResNet101 pre-trained on ImageNet dataset.
Note that we merely use the preview image to represent each furniture identity, as the instance patches are very various and noisy.
Then we perform clustering algorithm on the obtained feature set.
For simplicity, we use k-means in this paper, but other clustering approaches can also be used, e.g. AP\cite{frey2007clustering} and PIC\cite{lin2010power}.
To avoid identities of different categories clustered into one bucket, we employ a category supervised metric in k-means:
\begin{equation}
Dist(f_i, f_j)=
\begin{cases}
\|f_i - f_j\|_2 & y_i = y_j \\
+\infty & y_i \ne y_j
\end{cases}
\end{equation}
where $f_i, f_j$ denote the two feature vectors and $y_i, y_j$ denote the corresponding category code.
This metric guarantees that identities in different categories assigned to diverse attributes, while the items in one category are sufficiently distributed according to their visual appearances.
After obtaining the pseudo attributes, we fine-tune the network by classification loss.
Recent works on unsupervised learning, like DeepClustering (\cite{caron2018deep}), present the effectiveness of an iteration learning strategy between clustering and fine-tuning.
However, the iteration way is unable to improve the feature embedding in the context of this paper, which is detailed in the appendix.

\subsubsection{Pairwise verification subnet}

The pairwise verification subnet takes two feature vectors extracted from a paired furniture instance and identity as input.
They are then fused with element-wise subtraction and a ReLu.
After a FC layer, the output is a softmax layer with two nodes, representing whether or not the input pair is matched.
Besides this subtraction and binary cross-entropy loss, margin based contrastive loss is also widely used and a more sophisticated triplet loss is popular in person re-identification.
However, we empirically find that our simple method performs better in the furniture retrieval task.

A major caveat of the learning of the verification subnet is that the possible number of negative samples grows cubically with a large-scale training set.
After several epochs, most of the negative samples are relatively easy and contribute little for the loss.
In order to relieve this, we introduce online hard negative sample mining into the pipeline.
For each epoch, we pick the most dissimilar furniture identity for each furniture instance as the hardest negative samples.

The learning of the overall Siamese network is performed in a multi-task fashion.
Once the training is finished, we cast the backbone network as the embedding model to extract furniture features.
A ZCA whiten is also used to decorrelate the obtained feature.
Then furniture instance retrieval is performed by exhaustive Euclidean search over feature index of all the furniture identities.

\subsection{Furniture context embedding}

It is obvious that the furniture entities within one room are mutually related.
For example, chairs and tables usually simultaneously occur in one scene.
In order to mine these information, we train a neural embedding model with StarSpace\cite{wu2017starspace}, which is a strong library for efficient embedding learning.
In our case, one training example is an unordered set of furniture identities in a indoor scene design recorded in the real product.
We use more than 60M design data that covering 600K furniture identities to train an effective model, then we select the interested items of DeepFurniture dataset as the context embedding model in our pipeline.

\begin{algorithm}[!htb]\footnotesize
\renewcommand{\algorithmicrequire}{\textbf{Input:}}
\renewcommand\algorithmicensure {\textbf{Output:} }
\renewcommand\algorithmiccomment { }
\caption{The iterative context re-ranking algorithm}
\label{alg:re-rank}
\begin{algorithmic}[1]
\REQUIRE ~~\\
retrieval results $I$, retrieval distance $DF$, context lookup $C$, instance number $N$, top-k $K$\\
\ENSURE ~~\\
final results $I' = \{I_{1}',I_{2}',...,I_{n}'\}$ \\
\FOR{$j=1:K$}
\STATE Sort $I_{:j}$ by feature distance $DF_{:j}$
\STATE Add anchor $I_{1j}$ into $I'$ and remove it from $I$
\FOR{$i=2:N$ (\COMMENT{\small{\emph{Loop over instances}}})}
\STATE compute Eq. \ref{eq:dist} for each item in $I_{i:}$
\STATE add $\argmin_k {D_{ik}}$ into $I'$
\ENDFOR
\ENDFOR 
\end{algorithmic}
\end{algorithm}

In the end-to-end framework, we utilize the context embedding model to re-rank the retrieval results by measuring how good it is to arrange furniture identities together in one room.
After furniture instance detection and retrieval, a identity candidate set is obtained for
each instance.
The overall retrieval results can denoted as a matrix $I$, where each row corresponds to the 
top-k items searched for one instance : $I_{i} = (i_{i1},i_{i2},...,i_{ik})$, and the related feature distance matrix is denoted as $DF$.
In the re-ranking algorithm, it is complicated and time-consuming to get a global optimum, instead we adopt a simple and effective iterative method.
We denote the result matrix after re-ranking as $I'$.
To get the $i$-th ranked final set, \emph{i.e.} the $i$-th column in $I'$ , we first sort and choose the identity with the smallest visual feature distance from the $i$-th column of $I$ as an anchor.
Then we loop over the other instances and add the identity with the smallest incremental distance:
\begin{equation}\label{eq:dist}
     D_{ij} =  \alpha DF_{ij} + (1-\alpha) min_k Dist(C(I_{ij}), C(I'_{ik}))
\end{equation}
where $C$ presents the context embedding lookup and the second term is the minimized distance of context feature between the target item and the ones already added into the final set
By iteratively performing the above operation, we can finally get the re-ranked results.
The overall algorithm of the context re-ranking is summarized in Algorithm \ref{alg:re-rank}.

\section{Experiments and Results}
\label{sec5}

We demonstrate the effectiveness of our method by evaluating the performance on DeepFurniture dataset in multiple tasks including furniture detection, furniture instance retrieval and furniture set retrieval.
In this section, we first detail the experiment settings in Section 5.1.
Then we compare the improved Mask-RCNN with some object detection baselines in Section 5.2.
Section 5.3 shows the effectiveness of the attribute clustering, and the accuracy of instance retrieval is given in Section 5.4.
We finally report the performance of furniture set retrieval in Section 5.4.

\begin{table*}[]\small
\caption{Results of furniture detection and segmentation}
\label{tab:det} 
\begin{tabular}{c|c|ccc|ccc}
\hline
backbone & method & $AP_{box}$ & $AP_{box}^{IoU=0.50}$ & $AP_{box}^{IoU=0.75}$ & $AP_{mask}$ & $AP_{mask}^{IoU=0.50}$ & $AP_{mask}^{IoU=0.75}$ \\ 
\hline
ResNet50 & Faster R-CNN &  64.7  & 86.6& 75.3 & -- & -- & --  \\
ResNet50 & Mask R-CNN &   65.0 &  87.2& 74.6 & 51.5 & 74.5 & 53.3  \\
ResNet50 DCNv2 & Mask R-CNN & 68.2 & 88.2 & 78.6 & 54.0 & 77.6 &  56.0 \\
ResNet50 DCNv2 & Mask Score R-CNN & 68.6 & 88.4 & 79.4 & 55.5 & 79.2 & 57.4 \\
ResNet50 DCNv2 & \textbf{our} & \textbf{72.8} & \textbf{89.7} & \textbf{82.3} & \textbf{58.3} & \textbf{81.3} & \textbf{60.7} \\
\hline
ResNet101 & Faster R-CNN & 66.0 & 87.0 & 76.0 &  --  & -- & -- \\
ResNet101 & Mask R-CNN & 66.1 & 86.9 & 76.8 & 54.3 & 77.0 &56.1 \\ 
ResNet101 DCNv2 & Mask R-CNN & 70.6 &89.2 &79.7 & 56.6 & \textbf{79.9} &58.4 \\ 
ResNet101 DCNv2 & \textbf{our} & \textbf{73.0} & \textbf{90.1} & \textbf{81.2} & \textbf{56.6} & 78.3 &  \textbf{58.9} \\ 
\hline
\end{tabular}
\end{table*}

\subsection{Experiment settings}

The DeepFurniture dataset contains three types of examples, 
\emph{i.e.}, indoor images, furniture instances, furniture identities. 
For detection task, we samples $80\%$, about 19k indoor images, for training and the rest, nearly 5k images, for testing.
For the retrieval related tasks, it is more complicated to prepare the data.
First, $80\%$ identities are randomly selected into train set and the rest is cast as the testing set.
Then we extract furniture instances related to the training identities to train the Siamese network, and we generate 8k furniture instances as query set to evaluate the performance of furniture instance retrieval.
To evaluate the effectiveness of furniture context embedding and furniture set retrieval tasks, we randomly sample 3k indoor images that only contain furniture items in test set and meanwhile guarantee that these images have no overlap with the training images for detection.

We evaluate the improved Mask-RCNN models with the backbone of ResNetFPN50 and ResNetFPN101. Additional deformable convolution layers\cite{zhu2019deformable} are incorporated for better generalization ability.
Other settings remain the same as \cite{he2017mask}.
In the feature embedding network described in Section 4.2, we employ the backbone of ResNet101.
The number of attributes is fixed as $150$ in clustering, an we empirically set the relative ratio between the verification loss and classification loss to $10:1$.
In the inference pipeline, we utilize the fine-tuned network to extract feature embedding and perform ZCA whiten and $l_2$ normalization afterward.
In the re-ranking phase, the size of the context embedding model is set as $100$.

All the experiments are conducted on 2 GTX 1080Ti GPUs and all the deep learning models are implemented by PyTorch.
In particular, the improved Mask R-CNN is implemented based on the framework of maskrcnn-benchmark\cite{massa2018mrcnn}.
We develop the retrieval pipeline with the library of faiss\cite{johnson2019billion}.

\begin{figure}[tb]
\centering
\includegraphics[width=0.9\linewidth]{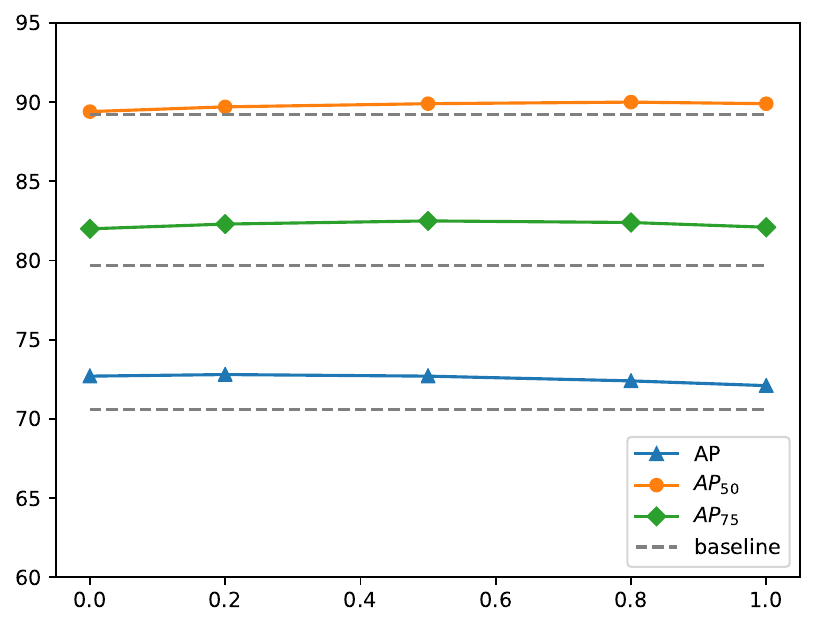}
\caption{The detection performance comparisons by varying the classification combining weight $\alpha$. The baseline refers to the performance of Mask R-CNN}
\label{fig:alpha}
\end{figure}


\subsection{Results of furniture detection and segmentation}

Table \ref{tab:det} displays the results of the proposed improved Mask R-CNN compared with baselines and related works.
It is shown that our detection result outperforms all the related methods in both ResNet-50 and ResNet-101 settings, and the segmentation performance is also improved for most cases.
In the inference of our model, the final detection score is obtained by combining the output scores of the two classifiers in the detection branch and mask branch.
We also analyze the influence of the combining weight $\alpha$ in Figure \ref{fig:alpha}, where $\alpha=1$ means only detection branch is used and $\alpha=0$ presents using the output of mask-based classifier as the final score.
Compared with baseline, our performance is still improved without using the mask branch in inference, because the introduced subnet can help regularize the model.
The best AP is achieved when $\alpha = 0.2$.

\subsection{Effectiveness of clustered attribute}

\begin{table}[htb]
\begin{center}
\caption{Ablation on the proposed Siamese model}
\label{tab:ablation}
\begin{tabular}{c|c|c|c}
\hline
classification & verification & MACC@1  & MACC@5  \\
\hline
-- & -- & 25.2 & 37.2  \\ 
category & -- & 21.4 & 34.3  \\ 
attribute & -- & 39.2 & 57.1  \\ 
-- & w/o ohnm  & 36.2 & 54.3  \\ 
attribute & w/o ohnm  & 46.4 & 65.3  \\ 
\hline
attribute & ohnm   & \textbf{52.1} & \textbf{71.0} \\
\hline
\end{tabular}
\end{center}
\end{table}

\begin{figure}[tb]
\centering
\includegraphics[width=0.9\linewidth]{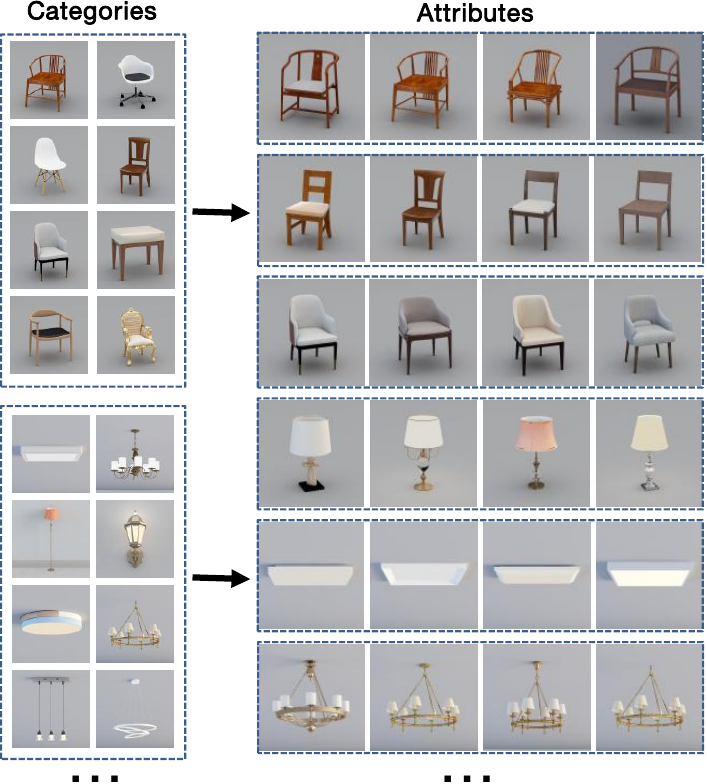}
\caption{Showcases of the self-clustered pseudo attributes}
\label{fig:kmeans}
\end{figure}

Figure \ref{fig:kmeans} shows some of the self-clustered pseudo attributes.
As category information is used as a supervision in the metric, the examples in one category are distributed into different attributes.
It is displayed that the items in one category are very diverse, but the examples in one attributes only present a specific visual appearance.
This can perform as a better regularization for image retrieval task.

We further analyze the quality of the attributes by evaluating the retrieval performance of the model fine-tuned by different supervision, as shown in Table \ref{tab:ablation}.
The baseline represents the ResNet101 pre-trained on ImageNet.
The model fine-tuned by the self-clustered attributes performs better than baseline and that fine-tuned by categories.
Note that the model fine-tuned by categories performs even worse than pre-trained model, because it give less visual distinction guidance for instance retrieval task.

\subsection{Results of feature instance retrieval}

\begin{table*}[tb]
\begin{center}
\caption{Furniture instance retrieval results}
\label{tab:instance-retrieval}
\begin{tabular}{|c|c|c|c|c|c|c|c|c|c|}
\hline
\multirow{2}*{Class} & \multicolumn{3}{|c|}{ACC@1} & \multicolumn{3}{|c|}{ACC@5} &
\multicolumn{3}{|c|}{ACC@10} \\
\cline{2-10} & baseline & Attribute & Siamese & baseline & Attribute & Siamese & baseline & Attribute & Siamese \\
\hline
cabinet/shelf & 18.5 & 35.5 & \textbf{54.6} & 29.6 & 54.0 & \textbf{74.0} & 34.3 & 61.1 & \textbf{79.2} \\
table & 8.9 & 21.2 & \textbf{33.5} & 16.4 & 35.2 & \textbf{57.1} & 19.3 & 40.2 & \textbf{54.4} \\
chair/stool & 18.3 & 27.5 & \textbf{46.0} & 28.1 & 44.6 & \textbf{67.3} & 33.1 & 52.6 & \textbf{74.3} \\
lamp & 30.2 & 40.4 & \textbf{61.0} & 48.6 & 63.9 & \textbf{85.2} & 54.4 & 72.8 & \textbf{89.3} \\
door & 16.4 & 20.5 & \textbf{33.0} & 26.8 & 36.9 & \textbf{51.2} & 30.7 & 44.4 & \textbf{57.7} \\
bed & 16.4 & 34.3 & \textbf{59.2} & 32.4 & 55.1 & \textbf{78.2} & 39.7 & 62.7 & \textbf{83.2} \\
sofa & 20.7 & 29.4 & \textbf{46.3} & 30.6 & 47.7 & \textbf{67.4} & 35.0 & 53.1 & \textbf{74.2} \\
plant & 28.8 & 39.1 & \textbf{69.8} & 43.6 & 66.2 & \textbf{88.0} & 50.6 & 74.0 & \textbf{91.6} \\
decoration & 75.5 & 83.4 & \textbf{90.8} & 43.6 & 66.2 & \textbf{88.0} & 50.6 & 74.0 & \textbf{91.6} \\
curtain & 3.9 & 16.9 & \textbf{29.8} & 9.0 & 29.5 & \textbf{48.0} & 16.6 & 37.1 & \textbf{54.5} \\
appliance & 40.0 & 43.4 & \textbf{48.6} & 57.7 & 61.1 & \textbf{77.7} & 62.3 & 68.0 & \textbf{82.3} \\
\hline
mean & 25.2 & 35.6 & \textbf{52.1} & 37.2 & 53.5 & \textbf{71.0} & 42.3 & 60.1 & \textbf{76.2} \\
\hline
\end{tabular}
\end{center}
\end{table*}

We first assess each component in the Siamese network by ablation study, as depicted in Table \ref{tab:ablation}.
The baseline represents the results of pre-trained ResNet101.
"ohnm" denotes the model trained with online hard negative mining for verification loss, while "w/o ohnm" denotes the model without adopting the trick.
This simple scheme improves the result by $5.7\%$.
It is displayed that only utilizing the verification loss decreases the result by more than $10\%$.
This demonstrates that the attributes guided classification subnet provide an effective regularization for the network training.
The best accuracy is achieved by mixing all, indicating the well complementary of the two branches.

\begin{figure}[tb]
\centering
\includegraphics[width=0.9\linewidth]{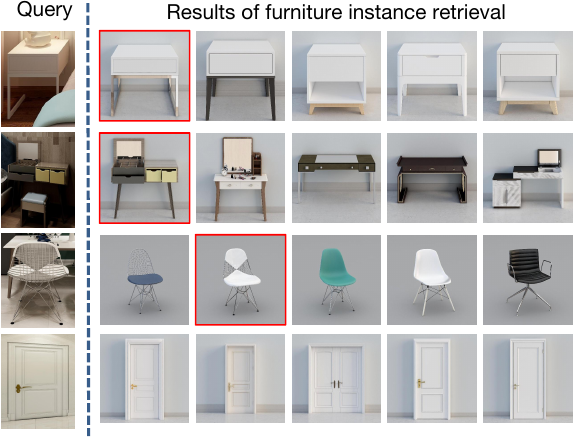}
\caption{Some examples of the furniture instance retrieval.}
\label{fig:query}
\end{figure}

Table \ref{tab:instance-retrieval} details the retrieval performance corresponding to each category in our dataset.
"Siamese" denotes the overall Siamese network, and "Attribute" presents the model fine-tuned by the pseudo attributes.
It is shown that the performance of each class is of high difference.
The performance of some classes like door and curtain, are relatively weak, because many furniture identities in these categories have very similar appearance.
Other mistakes may be caused by the diversity and occlusion of the furniture instances.
Some examples of the instance retrieval results are shown in Figure \ref{fig:query}, where the ground truth is indicated by red box.

\begin{table}[]
\begin{center}
\caption{Results of furniture set retrieval}
\label{tab:set}
\begin{tabular}{c|c|c|c}
\hline
detection & context   & ACC@1  & ACC@5 \\
\hline
gt & -- & 47.6\% & 61.9\% \\ 
gt  & \checkmark & 50.2\% & 63.3\% \\
our  & -- & 43.3\% & 56.1\% \\
our  & \checkmark  & 45.4\% & 57.4\%\\
\hline
& & SET ACC@5 & SET ACC@10 \\
\hline
gt & -- & 6.78\% & 9.64\% \\
gt & \checkmark & 9.53\% & 13.0\% \\
our & -- & 6.31\% & 8.35\% \\
our & \checkmark & 8.60\% & 11.3\% \\
\hline
\end{tabular}
\end{center}
\end{table}

\subsection{Results of furniture set retrieval}
We finally evaluate the end-to-end framework in Table \ref{tab:set} by the accuracy of furniture instance retrieval and furniture set retrieval.
Here, we use a test set containing 3k images and nearly 20k furniture instances and report both accuracy of instance retrieval as above experiments and set retrieval
The top-1 and top-5 instance accuracy of our system is $45.4\%$ and $57.4\%$ respectively, while the top-5 and top-10 set accuracy is $8.60\%$ and $11.3\%$ respectively.
It is noted that the accuracy of furniture set retrieval is relatively low because of the extremely challenging in the task.
Small mistakes of the detection and instance retrieval can be amplified to a big error in the final result.
In this table, we also show that the context embedding based re-ranking is effective to improve the performance in spite of its simplicity
It improves the feature set accuracy by $36.3\%$ relatively .
This demonstrates the potential relationship between furniture items is successfully learned by our context model.

\section{Conclusion}
\label{sec6}

In this paper, we present a large-scale dataset with rich annotations for furniture understanding and explore it for a new task, \emph{i.e.}, furniture set retrieval.
An end-to-end framework is proposed to solve this problem.
It contains three major modules: (1)improved Mask R-CNN for furniture detection (2) feature embedding network trained by integrating verification and classification loss (3) furniture context embedding for re-ranking.
Extensive results show the effectiveness of the three modules and the overall framework.
This system has already been utilized in several real products and we will release our dataset after paper acceptance.
Future works may contain the optimization of the re-ranking strategy and joint learning of the three modules.

{\small
\bibliographystyle{ieee_fullname}
\bibliography{main}

\begin{thebibliography}{10}\itemsep=-1pt

\bibitem{ahmed2015improved}
Ejaz Ahmed, Michael Jones, and Tim~K Marks.
\newblock An improved deep learning architecture for person re-identification.
\newblock In {\em CVPR}, pages 3908--3916, 2015.

\bibitem{badami20173d}
Ishrat Badami, Manu Tom, Markus Mathias, and Bastian Leibe.
\newblock 3d semantic segmentation of modular furniture using rjmcmc.
\newblock In {\em WACV}, pages 64--72. IEEE, 2017.

\bibitem{bai2016sparse}
Song Bai and Xiang Bai.
\newblock Sparse contextual activation for efficient visual re-ranking.
\newblock {\em IEEE Transactions on Image Processing}, 25(3):1056--1069, 2016.

\bibitem{bau2017network}
David Bau, Bolei Zhou, Aditya Khosla, Aude Oliva, and Antonio Torralba.
\newblock Network dissection: Quantifying interpretability of deep visual
  representations.
\newblock In {\em CVPR}, pages 6541--6549, 2017.

\bibitem{bodla2017soft}
Navaneeth Bodla, Bharat Singh, Rama Chellappa, and Larry~S Davis.
\newblock Soft-nms--improving object detection with one line of code.
\newblock In {\em ICCV}, pages 5561--5569, 2017.

\bibitem{caron2018deep}
Mathilde Caron, Piotr Bojanowski, Armand Joulin, and Matthijs Douze.
\newblock Deep clustering for unsupervised learning of visual features.
\newblock In {\em ECCV}, pages 132--149, 2018.

\bibitem{deng2009imagenet}
Jia Deng, Wei Dong, Richard Socher, Li-Jia Li, Kai Li, and Li Fei-Fei.
\newblock Imagenet: A large-scale hierarchical image database.
\newblock In {\em CVPR}, pages 248--255, 2009.

\bibitem{frey2007clustering}
Brendan~J Frey and Delbert Dueck.
\newblock Clustering by passing messages between data points.
\newblock {\em science}, 315(5814):972--976, 2007.

\bibitem{geng2016deep}
Mengyue Geng, Yaowei Wang, Tao Xiang, and Yonghong Tian.
\newblock Deep transfer learning for person re-identification.
\newblock {\em arXiv preprint arXiv:1611.05244}, 2016.

\bibitem{he2017mask}
Kaiming He, Georgia Gkioxari, Piotr Doll{\'a}r, and Ross Girshick.
\newblock Mask r-cnn.
\newblock In {\em ICCV}, pages 2961--2969, 2017.

\bibitem{he2016deep}
Kaiming He, Xiangyu Zhang, Shaoqing Ren, and Jian Sun.
\newblock Deep residual learning for image recognition.
\newblock In {\em CVPR}, pages 770--778, 2016.

\bibitem{hu2017visual}
Zhenhen Hu, Yonggang Wen, Luoqi Liu, Jianguo Jiang, Richang Hong, Meng Wang,
  and Shuicheng Yan.
\newblock Visual classification of furniture styles.
\newblock {\em ACM Transactions on Intelligent Systems and Technology (TIST)},
  8(5):67, 2017.

\bibitem{huang2019mask}
Zhaojin Huang, Lichao Huang, Yongchao Gong, Chang Huang, and Xinggang Wang.
\newblock Mask scoring r-cnn.
\newblock In {\em CVPR}, pages 6409--6418, 2019.

\bibitem{johnson2019billion}
Jeff Johnson, Matthijs Douze, and Herv{\'e} J{\'e}gou.
\newblock Billion-scale similarity search with gpus.
\newblock {\em IEEE Transactions on Big Data}, 2019.

\bibitem{krizhevsky2012imagenet}
Alex Krizhevsky, Ilya Sutskever, and Geoffrey~E Hinton.
\newblock Imagenet classification with deep convolutional neural networks.
\newblock In {\em NIPS}, pages 1097--1105, 2012.

\bibitem{lazebnik2006beyond}
Svetlana Lazebnik, Cordelia Schmid, and Jean Ponce.
\newblock Beyond bags of features: Spatial pyramid matching for recognizing
  natural scene categories.
\newblock In {\em CVPR}, pages 2169--2178, 2006.

\bibitem{li2018interiornet}
Wenbin Li, Sajad Saeedi, John McCormac, Ronald Clark, Dimos Tzoumanikas, Qing
  Ye, Yuzhong Huang, Rui Tang, and Stefan Leutenegger.
\newblock Interiornet: Mega-scale multi-sensor photo-realistic indoor scenes
  dataset.
\newblock {\em arXiv preprint arXiv:1809.00716}, 2018.

\bibitem{lim2013parsing}
Joseph~J Lim, Hamed Pirsiavash, and Antonio Torralba.
\newblock Parsing ikea objects: Fine pose estimation.
\newblock In {\em ICCV}, pages 2992--2999, 2013.

\bibitem{lin2010power}
Frank Lin and William~W. Cohen.
\newblock Power iteration clustering.
\newblock In {\em ICML}, pages 655--662, 2010.

\bibitem{fpn}
Tsung-Yi Lin, Piotr Doll{\'a}r, Ross Girshick, Kaiming He, Bharath Hariharan,
  and Serge Belongie.
\newblock Feature pyramid networks for object detection.
\newblock In {\em CVPR}, pages 2117--2125, 2017.

\bibitem{lin2017focal}
Tsung-Yi Lin, Priya Goyal, Ross Girshick, Kaiming He, and Piotr Doll{\'a}r.
\newblock Focal loss for dense object detection.
\newblock In {\em ICCV}, pages 2980--2988, 2017.

\bibitem{lin2014microsoft}
Tsung-Yi Lin, Michael Maire, Serge Belongie, James Hays, Pietro Perona, Deva
  Ramanan, Piotr Doll{\'a}r, and C~Lawrence Zitnick.
\newblock Microsoft coco: Common objects in context.
\newblock In {\em ECCV}, pages 740--755, 2014.

\bibitem{liu2017end}
Hao Liu, Jiashi Feng, Meibin Qi, Jianguo Jiang, and Shuicheng Yan.
\newblock End-to-end comparative attention networks for person
  re-identification.
\newblock {\em IEEE Transactions on Image Processing}, pages 3492--3506, 2017.

\bibitem{liu2016ssd}
Wei Liu, Dragomir Anguelov, Dumitru Erhan, Christian Szegedy, Scott Reed,
  Cheng-Yang Fu, and Alexander~C Berg.
\newblock Ssd: Single shot multibox detector.
\newblock In {\em ECCV}, pages 21--37, 2016.

\bibitem{massa2018mrcnn}
Francisco Massa and Ross Girshick.
\newblock {maskrcnn-benchmark: Fast, modular reference implementation of
  Instance Segmentation and Object Detection algorithms in PyTorch}.
\newblock \url{https://github.com/facebookresearch/maskrcnn-benchmark}, 2018.

\bibitem{radenovic2018fine}
Filip Radenovi{\'c}, Giorgos Tolias, and Ond{\v{r}}ej Chum.
\newblock Fine-tuning cnn image retrieval with no human annotation.
\newblock {\em IEEE transactions on pattern analysis and machine intelligence},
  41(7):1655--1668, 2018.

\bibitem{redmon2016you}
Joseph Redmon, Santosh Divvala, Ross Girshick, and Ali Farhadi.
\newblock You only look once: Unified, real-time object detection.
\newblock In {\em CVPR}, pages 779--788, 2016.

\bibitem{song2017semantic}
Shuran Song, Fisher Yu, Andy Zeng, Angel~X Chang, Manolis Savva, and Thomas
  Funkhouser.
\newblock Semantic scene completion from a single depth image.
\newblock In {\em CVPR}, pages 1746--1754, 2017.

\bibitem{fasterrcnn}
R.Girshick S.Ren, K.He and J.Sun.
\newblock Faster rcnn: Towards real-time object detection with region proposal
  networks.
\newblock In {\em NIPS}, pages 91--99, 2015.

\bibitem{DeepFace2014}
Yaniv Taigman, Ming Yang, Marc'Aurelio Ranzato, and Lior Wolf.
\newblock Deepface: Closing the gap to human-level performance in face
  verification.
\newblock In {\em CVPR}, pages 1701--1708, 2014.

\bibitem{wu2017starspace}
L. {Wu}, A. {Fisch}, S. {Chopra}, K. {Adams}, A. {Bordes}, and J. {Weston}.
\newblock Starspace: Embed all the things!
\newblock {\em arXiv preprint arXiv:{1709.03856}}, 2017.

\bibitem{xiao2018unified}
Tete Xiao, Yingcheng Liu, Bolei Zhou, Yuning Jiang, and Jian Sun.
\newblock Unified perceptual parsing for scene understanding.
\newblock In {\em ECCV}, pages 418--434, 2018.

\bibitem{zhao2017pspnet}
Hengshuang Zhao, Jianping Shi, Xiaojuan Qi, Xiaogang Wang, and Jiaya Jia.
\newblock Pyramid scene parsing network.
\newblock In {\em CVPR}, 2017.

\bibitem{zhong2017re}
Zhun Zhong, Liang Zheng, Donglin Cao, and Shaozi Li.
\newblock Re-ranking person re-identification with k-reciprocal encoding.
\newblock In {\em CVPR}, pages 1318--1327, 2017.

\bibitem{zhou2017scene}
Bolei Zhou, Hang Zhao, Xavier Puig, Sanja Fidler, Adela Barriuso, and Antonio
  Torralba.
\newblock Scene parsing through ade20k dataset.
\newblock In {\em CVPR}, pages 633--641, 2017.

\bibitem{zhu2019deformable}
Xizhou Zhu, Han Hu, Stephen Lin, and Jifeng Dai.
\newblock Deformable convnets v2: More deformable, better results.
\newblock In {\em CVPR}, pages 9308--9316, 2019.

\bibitem{zou2018layoutnet}
Chuhang Zou, Alex Colburn, Qi Shan, and Derek Hoiem.
\newblock Layoutnet: Reconstructing the 3d room layout from a single rgb image.
\newblock In {\em CVPR}, pages 2051--2059, 2018.

\end{thebibliography}
}

\clearpage
\appendix
\section{Analysis of deep clustering for attributes learning}
\label{app:a}

Inspired by DeepClustering\cite{caron2018deep}, we evaluate the method of iteratively updating the self-clustered attributes in the learning of the network.
That is, in each epoch, we first perform category supervised k-means to update the pesudo attributes and then fine-tune the network based on the latest attribute labels.
The performance on retrieval task is summarized in Table \ref{tab:deepcluster}.
It is shown than this scheme reduces the accuracy and the best performance is achieved by fixing the attributes at the beginning.
The reason may lie in that our model is fine-tuned on the pre-trained ResNet101 model and suffers over-fitting problem, while \cite{caron2018deep} trains the model from scratch.
Therefore, we fix the pesudo attributes in this paper.

\begin{table}[hb]
\begin{center}
\caption{Results of deep clustering scheme}
\label{tab:deepcluster}
\begin{tabular}{c|c|c|c}
\hline
epoch & macc@1 & macc@5 & macc@10 \\
\hline
pretrained & 27.52 & 39.80 & 44.7 \\
\textbf{1} & \textbf{39.2} & \textbf{57.1} & \textbf{0.6373} \\
2 & 34.4 & 52.7 & 60.0 \\
3 & 29.0 & 48.2 & 56.3\\
\hline
\end{tabular}
\end{center}
\end{table}

\section{Analysis of two structures in verification subnet}

\begin{figure}[b]
\centering
\includegraphics[width=1.0\linewidth]{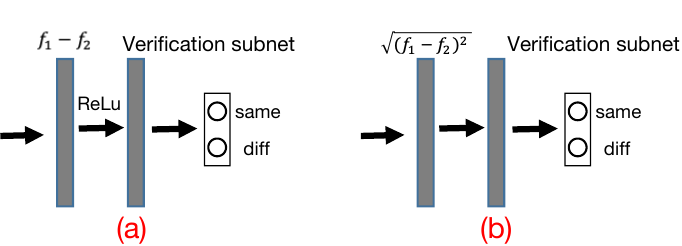}
\caption{We examine two structures in the verification subnet. (a) The input pair is fused with element-wise subtraction and a ReLu operator. (b) The input pair is fused with element-wise Euclidean distance operator. }
\label{fig:subnet}
\end{figure}

In our experiments, we compare two structures in the verification subnet shown in Figure \ref{fig:subnet}.
The structure (a) is the one used in this paper, where the input pair is fused with element-wise subtraction and a ReLu operator.
Different from that, the second one employs element-wise Euclidean distance operator which is the same as the metric used in retrieval pipeline.
In the same setting, the structure used in our model perform slightly better (macc@1 : $46.4\%$ vs. $44.6\%$).
This may be due to the sparsity introduced by the ReLu operator to relieve the over-fitting.

\section{Display of self-clustered pesudo attributes}

\begin{figure*}[htb]
\centering
\includegraphics[width=1.0\linewidth]{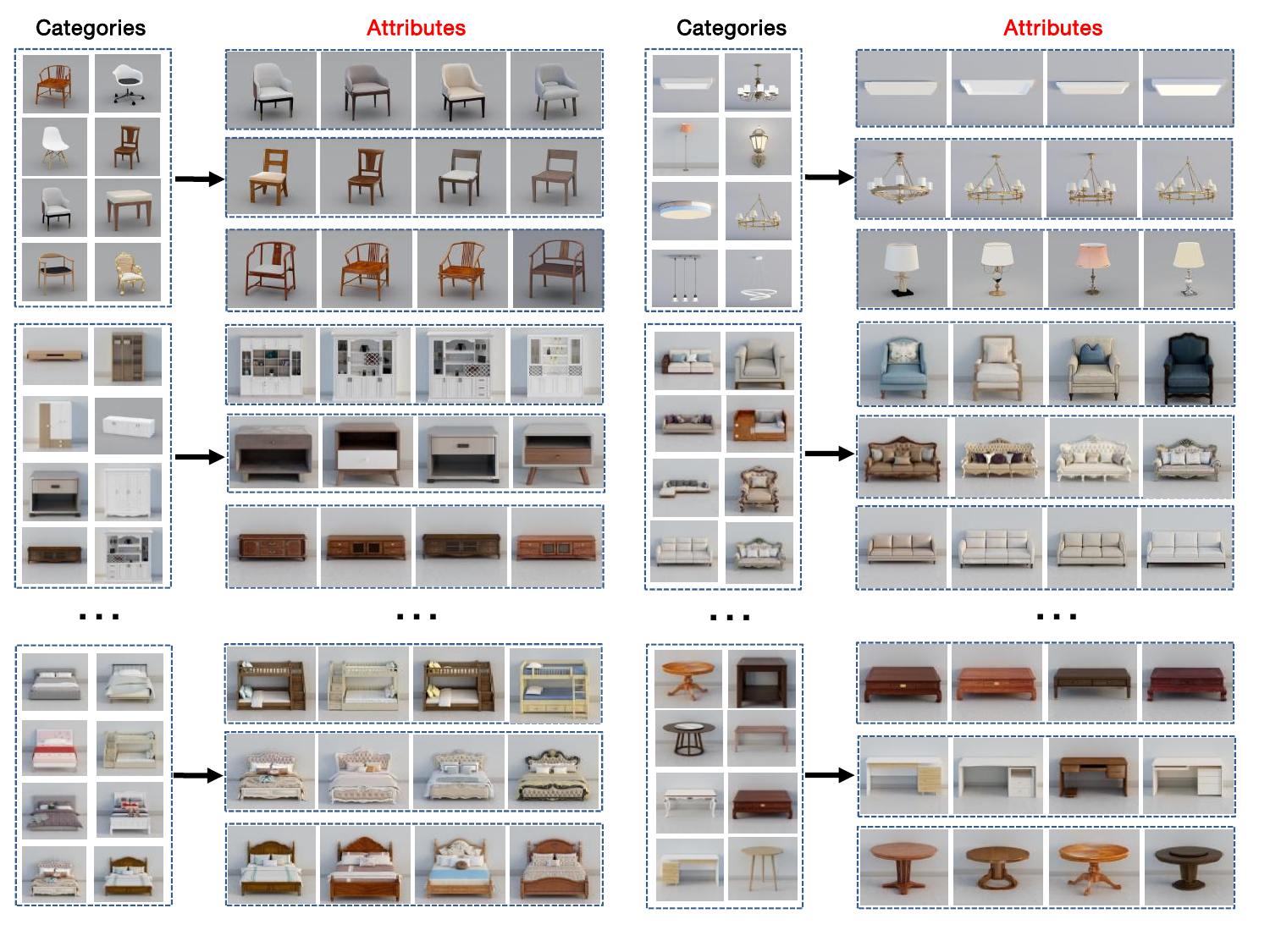}
\caption{Showcases of the self-clustered pesudo attributes. After clustering, the furniture identities in one category are divided into different attributes according to their visual appearance.}
\label{fig:attributes_more}
\end{figure*}

In Figure \ref{fig:attributes_more}, we show more cases of our pesudo attributes.
It is seen that these attributes are effective to assign the entities in each furniture category into different attributes base on their visual appearances.
This information is demonstrated to be effective to guide a network for the retrieval task.

\section{Showcases of our retrieval results}

\begin{figure*}[htb]
\centering
\includegraphics[width=1.0\linewidth]{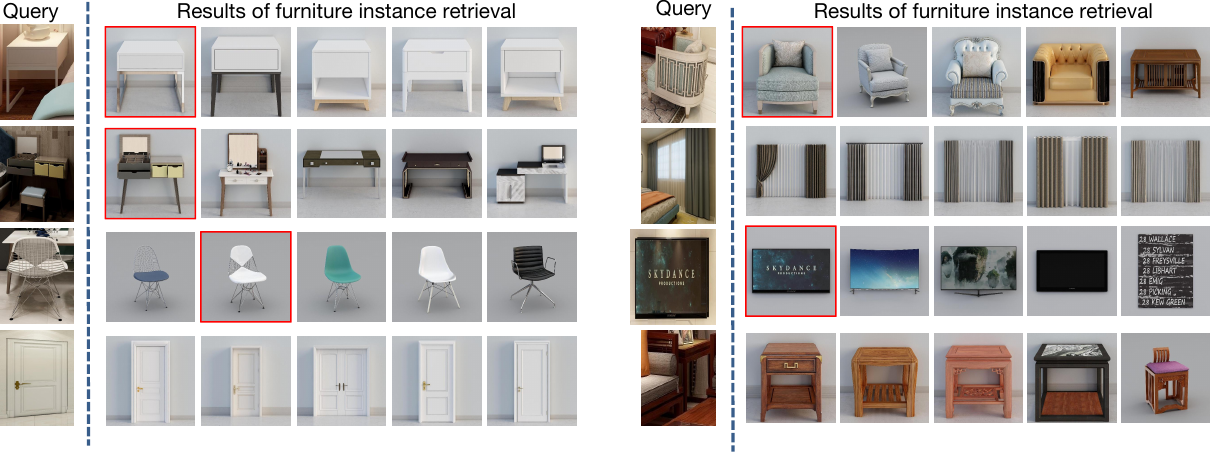}
\caption{Some example results of the furniture instance retrieval, where the ground truth is indicated by red box.}
\label{fig:more_query}
\end{figure*}

In Figure \ref{fig:more_query}, more results of the furniture instance retrieval are displayed.
It is shown that our model is robust to variance and occlusion of the query image to a large extent.
The performances of the door and curtain are relatively weak, because the entities in these categories are quite similar and only differ in some details.
Other mistakes may be caused by view and occlusion.

\begin{figure*}[htb]
\centering
\includegraphics[width=1.0\linewidth]{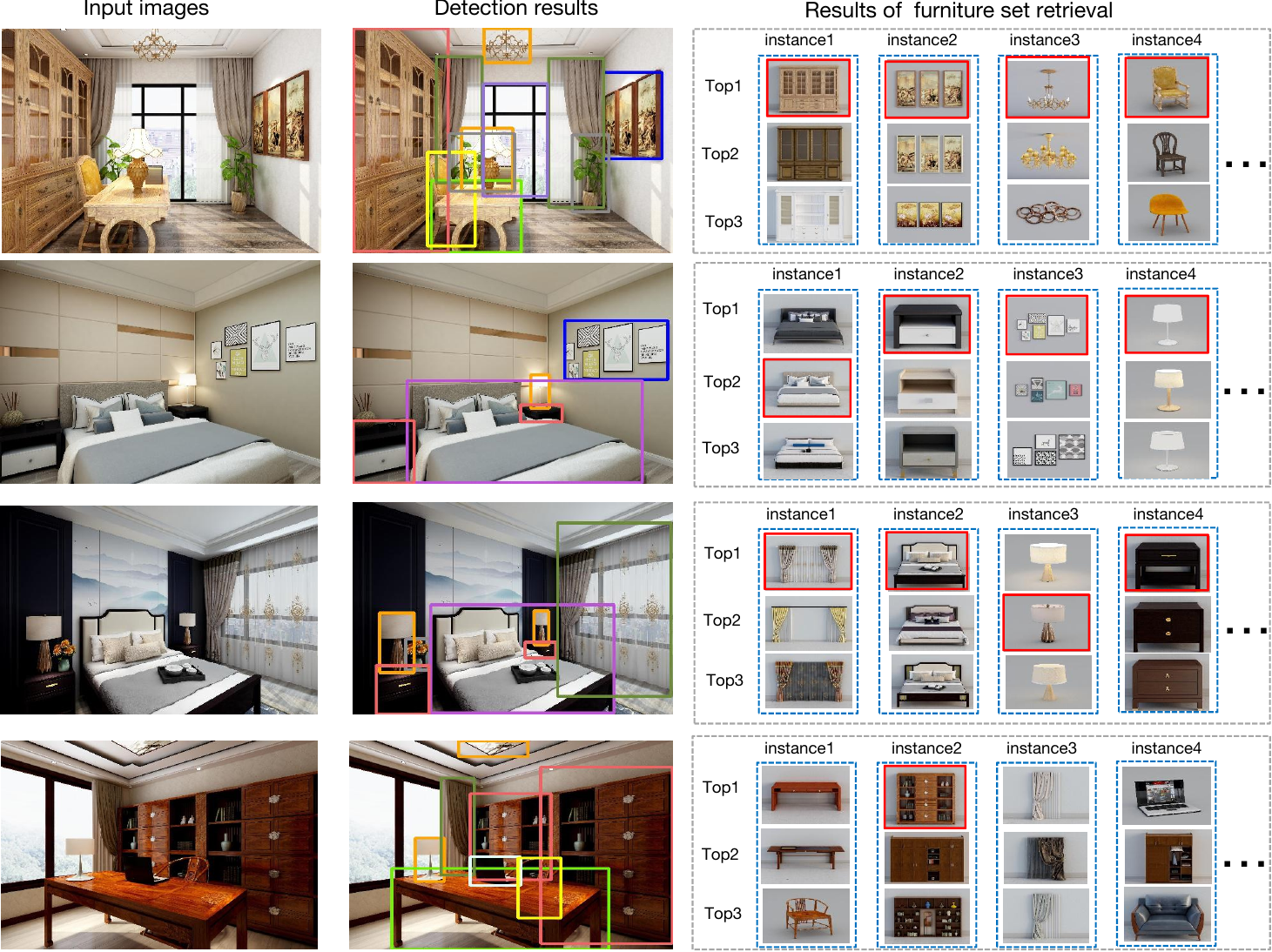}
\caption{Some example results of our end-to-end system, where the ground truth is indicated by red box.}
\label{fig:end}
\end{figure*}

In Figure \ref{fig:end}, we show some results of our end-to-end system.
Most major furniture items are successfully searched.
The set accuracy is mainly decreased by some extremely challenging categories like curtain and door, as the difference exists in very detail among the cases of these classes.
Even it is hard to obtain the exact furniture set, the search results are visually similar and acceptable for users.

\end{document}